\documentclass[letterpaper]{article}
\usepackage{aaai2027}
\nocopyright
\usepackage[hyphens]{url}
\usepackage{graphicx}
\usepackage{natbib}
\usepackage{caption}
\usepackage{amssymb}

\usepackage{booktabs}
\usepackage{array}
\usepackage{multirow}

\makeatletter
\let\aaaiOriginalFnsymbol\@fnsymbol
\renewcommand{\@fnsymbol}[1]{\ifcase#1\or\aaaiOriginalFnsymbol{2}\or\aaaiOriginalFnsymbol{1}\else\aaaiOriginalFnsymbol{#1}\fi}
\makeatother

\newcommand{\method}{\mbox{FIRM}}

\title{FIRM: Fine-Grained Intra-Token Representation of Masks for Remote Sensing Reasoning Segmentation}
\author{
Weidong Tang\textsuperscript{\rm 1, 3}\equalcontrib,
Kaiyu Li\textsuperscript{\rm 1}\equalcontrib\corresponding,
Yikai Wang\textsuperscript{\rm 2},
Yanan Wu\textsuperscript{\rm 3},\\
Haotian Gan\textsuperscript{\rm 4},
Shihong Wang\textsuperscript{\rm 1},
Xiangyong Cao\textsuperscript{\rm 1}\corresponding
}
\affiliations{
\textsuperscript{\rm 1}Xi'an Jiaotong University, Xi'an, China\\
\textsuperscript{\rm 2}Renmin University of China, Beijing, China\\
\textsuperscript{\rm 3}China Agricultural University, Beijing, China\\
\textsuperscript{\rm 4}Shaanxi University of Science and Technology, Xi'an, China
}

\begin{document}

\maketitle

\begin{abstract}
Reasoning segmentation requires multimodal large language models (MLLMs) to translate implicit instructions into precise pixel-level masks. MLLMs encode an image as visual tokens, each of which merges a group of image patches. In remote sensing images, small targets, thin structures, and adjacent instances can occupy different parts of the same visual token. Assigning a single binary mask label to such a token loses its internal spatial structure, causing nearby targets to merge and object boundaries to become coarse. To bridge this representational gap, we introduce FIRM, a Fine-grained Intra-token Representation of Masks. For each visual token, FIRM predicts a mask code that specifies an \(r\times r\) binary sub-cell pattern rather than a single foreground/background label. Given a target identified by the MLLM, the complete grid of mask codes is predicted in one mask pass.
Fixed lookup converts the predicted codes into a discrete sub-cell mask, while marginalizing the code distribution yields a soft structural field. To further recover fine-grained boundaries within each sub-cell, we introduce a lightweight continuous renderer that refines this field using pre-merge visual features and image details.
Across five reasoning and referring segmentation benchmarks on satellite and UAV images, FIRM achieves leading results, including \(70.5/80.5\) gIoU/cIoU on LaSeRS and a \(3.0\)-point average gain on EarthReason. These results demonstrate the value of explicitly representing intra-token mask patterns for fine-grained MLLM segmentation.
\end{abstract}

\begin{links}
\vspace{-4pt}
\link{Code}{https://github.com/earth-insights/FIRM}
\end{links}

\vspace{-12pt}
\section{Introduction}
\label{sec:introduction}

Recent multimodal large language models (MLLMs) have extended visual understanding from recognition and scene perception to fine-grained grounding and pixel-level localization \citep{bai2025qwen3vl,wang2024qwen2vl}. Reasoning segmentation requires an MLLM to infer a target from an implicit instruction by integrating world knowledge, object attributes, spatial relations, and contextual cues, and to delineate the target with a pixel-level mask \citep{lai2024lisa,jang2025mmr}. Compared with natural images, remote sensing images are typically characterized by overhead viewpoints, significant scale variations, densely distributed objects of the same category, and numerous small objects \citep{yuan2024rrsis,liu2024rotated,li2025segearth,xin2025segearth,ke2026pixdlm}. These characteristics not only increase the difficulty of target inference, but also impose greater demands on distinguishing adjacent instances, preserving fine-grained structures, and accurately delineating object boundaries.

\begin{figure}[t]
\centering
\includegraphics[width=\columnwidth]{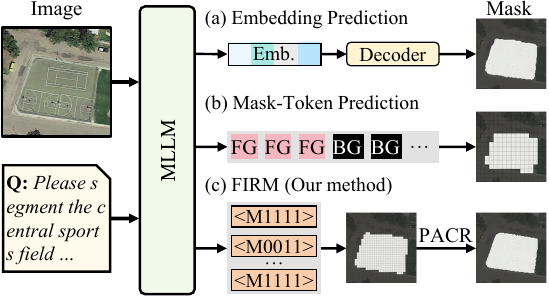}
\vspace{-20pt}
\caption{Comparison of MLLM-based segmentation paradigms.
(a) Embedding Prediction (e.g., SegEarth-R2 \citep{xin2025segearth}) relies on a separate mask decoder.
(b) Mask-Token Prediction (e.g., STAMP \citep{liu2026stamp})  represents each visual token with one foreground or background label.
(c) \method{} represents each visual token with an \(r\times r\) binary sub-cell pattern and refines the decoded mask with PACR.}
\label{fig:intro}
\vspace{-12pt}
\end{figure}

Existing MLLM segmentation methods largely follow two routes. The first passes an MLLM target representation through a segmentation token or mask query to SAM or a separate mask decoder, as shown in Figure~\ref{fig:intro}(a). This route includes LISA, GLaMM, PixelLM, and PSALM \citep{lai2024lisa,rasheed2024glamm,ren2024pixellm,zhang2024psalm,zhang2024omg,kirillov2023segment}, as well as the remote sensing methods SegEarth-R1 and SegEarth-R2 \citep{li2025segearth,xin2025segearth}. Although a specialist decoder can exploit high resolution visual features, it introduces a separate module for mask generation and places the mask outside the MLLM output interface. The second route represents a mask directly with outputs generated by the MLLM, including coordinates, polygons, or mask tokens \citep{liu2023polyformer,lan2024text4seg,wang2025himtok}. This route makes mask prediction part of the unified output interface, while raising a representational question for dense masks. In most MLLMs, the connector merges neighboring image patches into one visual token. When the mask within such a token is represented by a single binary label, different foreground patterns across the merged patches cannot be distinguished. This loss of local structure is particularly severe when a target occupies only part of a token or shares it with a nearby instance.

To address this gap, we propose \method{}, a Fine-grained Intra-token Representation of Masks aligned with the visual compression structure of MLLMs. Rather than assigning one foreground/background label to each visual token, \method{} predicts a mask code that specifies an \(r\times r\) binary sub-cell pattern. The set of mask codes covers every possible pattern at this granularity, allowing local foreground layouts to be represented explicitly. Across an \(H\times W\) grid of visual tokens, these patterns form an \(rH\times rW\) sub-cell mask without changing the number of token aligned predictions. Once the MLLM identifies a target, the complete grid of mask codes is predicted in one parallel mask pass. A fixed lookup table converts the predicted codes into a discrete sub-cell mask, while marginalizing the code distributions produces a soft structural field. To further recover fine-grained boundaries within each sub-cell, we introduce a lightweight continuous renderer. It refines the structural field using visual features before patch merging and shallow image details, while keeping the predicted mask pattern as its reference. In this way, \method{} progressively maps visual tokens to sub-cell patterns and then to continuous pixel-level masks.

In summary, our contributions are as follows:
\begin{itemize}
    \item We formulate the loss of local mask structure within compressed visual tokens as an intra-token representation problem. This perspective explains why a single binary mask label cannot describe small targets and adjacent instances that occupy the same visual token.

    \item We introduce \method{}, which represents each visual token with a mask code specifying an \(r\times r\) binary sub-cell pattern. Fixed lookup and code marginalization produce a discrete sub-cell mask and a soft structural field, while a lightweight continuous renderer further recovers boundary details within each sub-cell.

    \item We evaluate \method{} on LaSeRS, EarthReason, DRSeg, RRSIS-D, and RISBench, covering reasoning and referring segmentation in satellite and UAV images. \method{} achieves leading results, and the granularity study shows that intra-token mask patterns and continuous rendering provide complementary improvements with minimal inference overhead.
\end{itemize}

\section{Related Work}
\label{sec:related}

\paragraph{Reasoning segmentation.}
Referring segmentation grounds targets that are explicitly described in language. Existing methods improve this grounding through cross-modal fusion \citep{yang2022lavt}, generalized target handling \citep{hu2023beyond,liu2023gres}, and polygon-based decoding \citep{liu2023polyformer}. Reasoning segmentation further requires the model to infer the intended target from knowledge, attributes, or spatial relations before producing its mask \citep{lai2024lisa}. Several methods use explicit reasoning chains or reinforcement learning to improve target discovery and grounding \citep{bao2024cores,liu2025segzero,liu2025visionreasoner,lu2025rsvp,you2025seg}. More recent formulations further develop reward design and rollout strategies for this reasoning process \citep{zhu2026lens,he2026dr2seg,sun2026drseg}. These advances strengthen semantic reasoning and target identification.

\paragraph{Output interfaces for MLLM segmentation.}
MLLM segmentation methods connect language reasoning to pixel prediction through different output interfaces. One common interface passes the hidden states of generated segmentation tokens to a separate mask decoder \citep{lai2024lisa,ren2024pixellm,rasheed2024glamm,zhang2024groundhog,zhang2024omg,zhang2024psalm,wei2025instructseg}, often using a promptable foundation model for pixel-level decoding \citep{kirillov2023segment,yuan2025sa2va}. Another brings mask representation more directly into the MLLM output interface. Some methods generate coordinates, polygons, or annotator trajectories \citep{wang2023visionllm,liu2023polyformer,zhu2025segagent}, while others use textual patch labels or learned mask tokens \citep{lan2024text4seg,lan2025text4segpp,deng2026llamaseg,wang2025himtok,wang2025alto}. UFO predicts multiple mask token embeddings and matches them with visual features to obtain finer masks \citep{tang2025ufo}, while SELF1E restores visual features before patch merging and matches them with a single target embedding \citep{zhang2026self1e}. STAMP predicts a complete grid of binary mask tokens in one bidirectional mask pass after target generation \citep{liu2026stamp}. These methods mainly differ in how masks are decoded or generated, while the local mask structure represented at each compressed visual token remains less explored.

\begin{figure*}[t]
\centering
\includegraphics[width=\linewidth]{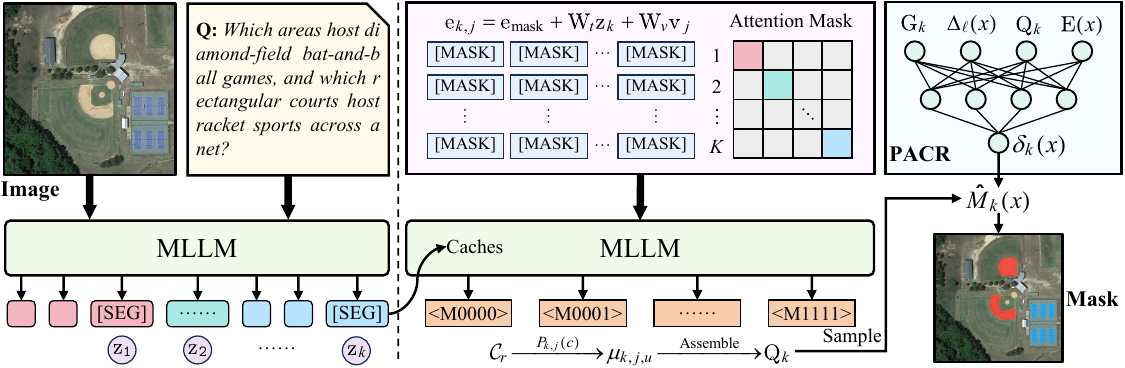}
\caption{Overview of the \method{} pipeline. The MLLM first identifies the targets and emits one \texttt{[SEG]} token for each target. Conditioned on the corresponding target representation, a grid of \texttt{[MASK]} queries predicts a distribution over mask codes at every visual-token position in one mask pass, with attention blocked across targets. Fixed lookup decodes the most probable codes into a discrete sub-cell mask, while marginalization yields a soft structural field. PACR further refines this field into the final pixel-level mask.}
\label{fig:pipeline-overview}
\end{figure*}

\paragraph{Language-guided segmentation in remote sensing.}
Language-guided segmentation in remote sensing images has progressed from dedicated referring segmentation datasets \citep{yuan2024rrsis} and specialized architectures \citep{liu2024rotated,dong2024cross} to larger benchmarks and adaptations of foundation models \citep{yang2025large,chen2025rsrefseg,yao2025remotesam}. MLLM-based systems further support pixel-level conversation and grounding in geospatial images \citep{ou2025geopix,shabbir2025geopixel,quenum2025lisat,zhou2024geoground}. Recent studies extend this direction to implicit geospatial reasoning \citep{li2025segearth,yao2025remotereasoner,zhang2025think2seg}, broader instruction types and segmentation granularities \citep{xin2025segearth,ni2025unigeoseg}, reasoning segmentation in UAV images \citep{ke2026pixdlm}, and textual mask representations for high resolution outputs \citep{tiwary2026magseg}. These advances support semantically richer instructions and more spatially precise outputs. \method{} complements this progress by explicitly representing local mask patterns within compressed visual tokens, which is particularly useful for small, thin, and closely spaced targets in remote sensing images.

\section{Method}
\label{sec:method}

\subsection{Problem Formulation and Overview}

Let \(\mathbf I\in\mathbb R^{H_I\times W_I\times 3}\) be an image and \(q\) a language instruction. The task is to infer the \(K\) targets specified by \(q\) and predict a pixel-level binary mask \(\mathbf M_k\in\{0,1\}^{H_I\times W_I}\) for each target \(k\). The vision encoder divides \(\mathbf I\) into patches of side \(p\), and the connector merges each \(m\times m\) group of patch features into one visual token, producing an \(H\times W\) grid of \(N=HW\) visual tokens \(\{\mathbf v_j\}_{j=1}^{N}\). Each visual token is aligned with an \(mp\times mp\) image region. One foreground or background label per visual token cannot preserve spatial structure within the corresponding image region. \method{} instead predicts a target-specific mask code \(c_{k,j}\in\mathcal C_r\) at each visual-token position. Each code specifies an \(r\times r\) binary sub-cell pattern, so decoding the complete \(H\times W\) code grid produces an \(rH\times rW\) discrete mask. The subdivision granularity \(r\) is independent of the connector merge factor \(m\). Figure~\ref{fig:pipeline-overview} summarizes the full pipeline. After the MLLM identifies target \(k\), the hidden state \(\mathbf z_k\) of its \texttt{[SEG]} token conditions prediction of the complete mask-code grid.

\subsection{Fine-Grained Intra-Token Mask Representation}

For a subdivision granularity \(r\), we introduce a set of \(2^{r^2}\) mask codes, each represented by a registered token in the language-model vocabulary,
\begin{equation}
\mathcal C_r=\{0,\ldots,2^{r^2}-1\},\qquad
c\longleftrightarrow\mathbf b(c)\in\{0,1\}^{r^2},
\label{eq:codebook}
\end{equation}
where the fixed bijection \(\mathbf b\) maps each code \(c\) to an \(r\times r\) binary sub-cell pattern in row-major order. The element \(b_u(c)\) specifies whether sub-cell \(u\) is foreground. A single mask code thus jointly represents the foreground and background states of all \(r^2\) sub-cells associated with one visual token.

To construct the supervision for target \(k\), we divide its ground-truth mask \(\mathbf M_k\) into an \(rH\times rW\) grid of sub-cells. The average mask occupancy within each sub-cell is thresholded at \(\tau\), producing
\(\mathbf Y_k\in\{0,1\}^{rH\times rW}\). For each visual-token position \(j\), the \(r\times r\) block in \(\mathbf Y_k\) is flattened in row-major order and mapped through \(\mathbf b^{-1}\) to obtain the ground-truth mask code \(c^\star_{k,j}\). During decoding, the fixed lookup table retrieves \(\mathbf b(c)\) for each predicted code and places the resulting pattern at its corresponding position. Assembling all decoded patterns produces an \(rH\times rW\) discrete sub-cell mask. Both encoding and decoding use fixed mappings and spatial rearrangement without a learned mask decoder.

The mask-code set contains every possible binary pattern over the \(r^2\) sub-cells. Increasing \(r\) therefore refines the discrete mask while retaining \(H\times W\) code predictions, although the number of mask codes grows as \(2^{r^2}\). The decoded mask remains constant within each sub-cell, and the continuous renderer subsequently recovers finer boundary details inside these regions.

\subsection{Target-Conditioned Mask Prediction}

After the autoregressive reasoning pass identifies a target, the hidden state \(\mathbf z_k\) of its \texttt{[SEG]} token serves as the representation of target \(k\). For target \(k\), \method{} creates a grid of \(N\) \texttt{[MASK]} queries, one for each visual-token position. The input embedding of the query at position \(j\) is
\begin{equation}
\mathbf e_{k,j}
=
\mathbf e_{\rm mask}
+
\mathbf W_t\mathbf z_k
+
\mathbf W_v\mathbf v_j ,
\label{eq:placeholder}
\end{equation}
where \(\mathbf e_{\rm mask}\) is the learned embedding of the registered special token \texttt{[MASK]}, and \(\mathbf W_t\) and \(\mathbf W_v\) project the target representation and visual token \(\mathbf v_j\), respectively. Each query therefore combines the semantics of target \(k\) with the visual content at position \(j\). It is assigned the multimodal positional index of \(\mathbf v_j\), requiring no additional positional embedding.

During the mask pass, the original multimodal prefix retains causal attention and does not attend to the added mask queries. Each \texttt{[MASK]} query attends to the full prefix and to all mask queries associated with the same target, whereas attention between queries of different targets is blocked. The queries for one target can therefore exchange information across the complete visual-token grid while remaining separated from those of other targets.

The mask pass returns one hidden state \(\mathbf h_{k,j}\) for each query. Restricting the language-model output head to the registered mask-code tokens gives a distribution over \(\mathcal C_r\),
\begin{equation}
P_{k,j}(c)
=
{\rm softmax}_{c\in\mathcal C_r}
\left(\mathbf w_c^{\mathsf T}\mathbf h_{k,j}\right),
\label{eq:posterior}
\end{equation}
where \(\mathbf w_c\) is the output head for mask code \(c\). The probability \(P_{k,j}(c)\) specifies how likely the visual-token position \(j\) is to take mask code \(c\) for target \(k\). The distributions \(\{P_{k,j}\}_{j=1}^{N}\) form the structural posterior of target \(k\), from which the discrete sub-cell mask and soft structural field are derived.

\subsection{Discrete Readout and Structural Field}

For a discrete mask, we select the most likely mask code at each visual-token position, \(\hat c_{k,j}=\arg\max_{c}P_{k,j}(c)\).
The fixed lookup table in Equation~(\ref{eq:codebook}) decodes each \(\hat c_{k,j}\) into its binary sub-cell pattern. Assembling the decoded patterns over all visual-token positions produces the discrete mask
\(\hat{\mathbf Y}_k\in\{0,1\}^{rH\times rW}\).
The distribution over mask codes also yields a soft structural field. For sub-cell \(u\) within visual token \(j\), we marginalize over all codes,
\begin{equation}
\mu_{k,j,u}=\sum_{c\in\mathcal C_r}P_{k,j}(c)\,b_u(c),
\label{eq:marginal}
\end{equation}
where \(\mu_{k,j,u}\) is the foreground probability of sub-cell \(u\). Placing each \(\mu_{k,j,u}\) at the corresponding sub-cell position yields the soft structural field
\begin{equation}
\mathbf Q_k={\rm Assemble}\!\left(\{\mu_{k,j,u}\}_{j,u}\right)\in[0,1]^{rH\times rW}.
\label{eq:assemble}
\end{equation}
The discrete mask \(\hat{\mathbf Y}_k\) and soft field \(\mathbf Q_k\) are thus complementary readouts of the same distribution over mask codes. Because the marginalization and assembly are differentiable, \(\mathbf Q_k\) provides a differentiable structural reference for continuous rendering.

\subsection{Posterior-Anchored Continuous Rendering}

Although the structural field \(\mathbf Q_k\) provides foreground probabilities for individual sub-cells, it is defined only on the \(rH\times rW\) sub-cell grid. To recover boundary details within each sub-cell, the Posterior-Anchored Continuous Renderer (PACR) predicts the mask on a denser grid of \(\kappa rH\times\kappa rW\) query points. At each query coordinate, PACR uses the structural field as its base prediction and learns a residual correction from finer visual evidence.

\paragraph{Target-conditioned fusion.}

The mask-query hidden states \(\{\mathbf h_{k,j}\}_{j=1}^{N}\) are linearly projected and rearranged by pixel shuffle from the \(H\times W\) visual-token grid to the \(rH\times rW\) sub-cell grid. The resulting features are combined with the pre-merge visual tensor \(\mathbf F^{\rm pre}\), the spatially broadcast target representation \(\mathbf z_k\), and the channel-broadcast structural field \(\mathbf Q_k\). The inputs are projected to a common channel dimension, and \(\mathbf F^{\rm pre}\) is resized to the sub-cell grid when its spatial resolution differs. Two local mixer blocks fuse their sum into a target-conditioned feature \(\mathbf G_k\). A shallow stem separately extracts an image-detail feature \(\mathbf E\) from the RGB image resized to the query-grid resolution. In this fusion, \(\mathbf Q_k\) provides the predicted target structure, while the visual features provide local evidence for refining its boundaries.

\paragraph{Posterior-anchored residual.}
For a query point \(x\), let \(\mathcal N(x)\) denote its four neighboring points on the sub-cell grid. We use \(\omega_\ell(x)\) for the corresponding bilinear weight and \(\Delta_\ell(x)\) for the relative offset from \(x\) to neighbor \(\ell\). For each neighbor, the renderer concatenates the fused feature \(\mathbf G_k(\ell)\), relative offset \(\Delta_\ell(x)\), structural-field value sampled at \(x\), and image-detail feature \(\mathbf E(x)\) into \(\mathbf a_{k,\ell}(x)\). A shared pointwise network \(\phi_\theta\) predicts a scalar correction from each neighboring point, giving
\begin{equation}
\delta_k(x)
=
\sum_{\ell\in\mathcal N(x)}
\omega_\ell(x)\,
\phi_\theta\!\left(\mathbf a_{k,\ell}(x)\right).
\label{eq:residual}
\end{equation}
The final foreground probability is obtained by adding this correction to the sampled structural logit,
\begin{equation}
\hat M_k(x)
=
\sigma\!\left(
{\rm Sample}\!\left({\rm logit}(\mathbf Q_k),x\right)
+
\delta_k(x)
\right),
\label{eq:render}
\end{equation}
where \({\rm Sample}\) denotes bilinear sampling, \({\rm logit}(\mathbf Q_k)\) is clamped for numerical stability, and \(\sigma\) is the logistic function. The last layer of \(\phi_\theta\) is initialized to zero, making \(\delta_k(x)=0\) at the beginning of training. The initial prediction is therefore determined entirely by the structural field, after which PACR learns residual corrections that refine local boundaries while retaining the predicted sub-cell structure.

\begin{table*}[t]
\centering
\setlength{\tabcolsep}{0pt}
{\small
\begin{tabular}{@{}>{\raggedright\arraybackslash}p{2.50cm}*{10}{>{\centering\arraybackslash}p{1.52cm}}@{}}
\toprule
\multirow{2}{*}{Method} & \multicolumn{3}{c}{Segmentation Granularity} & \multicolumn{2}{c}{Target Multiplicity} & \multicolumn{2}{c}{Reasoning Requirement} & \multicolumn{2}{c}{Expression Length} & \multirow{2}{*}{Avg.} \\
\cmidrule(lr){2-4} \cmidrule(lr){5-6} \cmidrule(lr){7-8} \cmidrule(lr){9-10}
 & Semantic & Instance & Part & Single & Multiple & Explicit & Implicit & Short & Long & \\
\midrule
\multicolumn{11}{@{}l}{\textit{With Mask Decoder}} \\
LISA-7B & 26.4/23.2 & 20.5/25.0 & 16.1/11.6 & 37.3/32.2 & 18.2/22.4 & 27.1/24.3 & 21.5/25.6 & 34.1/27.8 & 38.4/33.9 & 26.6/25.1 \\
LISA-13B & 27.0/24.5 & 22.3/25.6 & 17.7/13.1 & 38.4/34.2 & 19.9/23.5 & 27.1/25.5 & 22.6/25.8 & 35.2/28.0 & 38.4/34.3 & 27.6/26.1 \\
PixelLM-7B & 32.0/32.8 & 26.6/30.0 & 13.2/16.5 & 44.3/40.4 & 20.2/23.5 & 25.0/23.1 & 23.9/21.9 & 41.6/38.9 & 37.1/34.5 & 29.3/29.1 \\
PixelLM-13B & 31.6/34.0 & 27.5/30.2 & 15.8/17.6 & 42.2/40.5 & 20.9/22.4 & 26.3/24.4 & 25.9/22.1 & 42.0/39.1 & 37.1/34.5 & 29.9/29.4 \\
GLaMM-ft-7B & 44.8/47.9 & 41.2/48.3 & 32.6/42.7 & 47.3/50.3 & 32.2/41.0 & 59.1/60.3 & 42.6/44.8 & 50.4/54.8 & 42.6/44.8 & 43.6/48.3 \\
\(M^2SA\)-7B & 30.1/33.0 & 23.0/24.8 & 18.6/17.2 & 45.4/37.6 & 20.9/24.8 & 35.8/30.4 & 23.3/26.7 & 35.8/32.8 & 41.5/36.7 & 30.5/29.3 \\
GeoPixel-8B & 51.4/57.2 & 44.1/49.3 & 43.9/52.4 & 55.0/45.8 & \underline{49.2}/49.7 & 66.5/61.3 & 41.1/58.3 & 51.1/59.3 & 51.4/63.2 & 50.4/55.2 \\
SegEarth-R2-3B & 60.2/71.8 & 65.4/70.3 & 64.8/68.3 & 55.1/\underline{69.2} & 38.3/56.2 & 78.4/80.4 & 42.8/59.7 & 60.2/69.9 & 50.1/65.7 & 57.2/67.9 \\
\midrule
\multicolumn{11}{@{}l}{\textit{Mask-Decoder-Free}} \\
Text4Seg-4B & 44.0/40.7 & 59.2/62.1 & 65.6/62.3 & 47.5/47.7 & 34.5/47.6 & 70.5/76.1 & 44.8/62.4 & 53.7/52.1 & 53.3/54.9 & 52.6/56.2 \\
UFO-8B & 45.5/54.5 & 56.2/60.9 & 54.4/69.1 & 49.9/52.8 & 37.6/54.7 & 61.7/72.9 & 43.6/56.8 & 53.2/64.8 & 53.8/64.6 & 50.6/61.2 \\
SELF1E-2B & 61.3/71.6 & 65.0/70.0 & 65.3/75.6 & 62.2/63.5 & 47.5/68.2 & 73.9/78.3 & 56.5/74.8 & 69.5/78.3 & 68.3/78.1 & 63.3/73.1 \\
SELF1E-8B & \underline{63.7}/\underline{73.5} & \underline{73.7}/\underline{74.7} & \underline{73.4}/\underline{78.1} & \underline{65.1}/65.5 & \textbf{52.1}/\underline{68.8} & \underline{80.4}/\underline{87.6} & \underline{61.7}/\underline{79.0} & \underline{74.6}/\underline{83.8} & \textbf{74.0}/\underline{82.7} & \underline{68.7}/\underline{77.1} \\
STAMP-2B & 38.2/55.9 & 51.3/63.0 & 53.4/73.5 & 46.2/57.4 & 30.7/56.4 & 59.8/74.8 & 40.3/62.8 & 51.0/67.2 & 49.5/67.1 & 46.7/64.2 \\
STAMP-7B & 41.1/56.9 & 56.7/69.4 & 60.7/75.5 & 46.2/55.7 & 32.6/55.0 & 65.7/77.2 & 42.5/67.1 & 52.7/66.6 & 49.7/67.6 & 49.8/65.7 \\
\textbf{FIRM-4B} & \textbf{67.2}/\textbf{77.9} & \textbf{77.3}/\textbf{77.2} & \textbf{79.1}/\textbf{83.8} & \textbf{66.4}/\textbf{71.4} & 47.8/\textbf{75.4} & \textbf{84.5}/\textbf{90.5} & \textbf{64.6}/\textbf{79.6} & \textbf{75.2}/\textbf{85.2} & \underline{72.1}/\textbf{83.3} & \textbf{70.5}/\textbf{80.5} \\
\bottomrule
\end{tabular}
}
\caption{Comparison on LaSeRS. Each entry is gIoU/cIoU (\%); bold and underline mark the best and second-best value.}
\label{tab:lasers}
\end{table*}

\subsection{Learning Objective and Inference}

Let \(\mathcal J\) denote the set of supervised target-position pairs \((k,j)\) in a training batch. Each pair has a predicted distribution \(P_{k,j}\) and a ground-truth mask code \(c^\star_{k,j}\). We train the intra-token mask representation using
\begin{equation}
\mathcal L_{\rm rep}
=
\mathcal L_{\rm CE}
+
\mathcal L_{\rm Ham}
+
\mathcal L_{\rm Dice}^{\rm str}.
\label{eq:lrep}
\end{equation}
The weighted categorical loss supervises the identity of each mask code,
\begin{equation}
\mathcal L_{\rm CE}
=
-\frac{1}{|\mathcal J|}
\sum_{(k,j)\in\mathcal J}
w(c^\star_{k,j})
\log P_{k,j}(c^\star_{k,j}),
\label{eq:lce}
\end{equation}
where \(w(\cdot)\) downweights the all-background code to reduce the foreground-background imbalance over the complete visual-token grid.
Categorical cross-entropy does not encode the structural similarity between different mask codes. We therefore introduce an expected Hamming loss,
{\small
\begin{equation}
\mathcal L_{\rm Ham}
=
\frac{1}{|\mathcal J|}
\sum_{(k,j)\in\mathcal J}
w(c^\star_{k,j})
\sum_{c\in\mathcal C_r}
P_{k,j}(c)\,
d_H(c,c^\star_{k,j}),
\label{eq:lham}
\end{equation}
}where \(d_H(c,c^\star_{k,j})\) counts the sub-cells on which the two codes disagree. This loss penalizes a predicted code according to the number of incorrect sub-cell labels in its binary pattern. In addition, \(\mathcal L_{\rm Dice}^{\rm str}\) is computed between the soft structural field \(\mathbf Q_k\) and the sub-cell supervision \(\mathbf Y_k\), averaged over targets. It directly supervises the spatial overlap of the assembled sub-cell mask.

The rendered mask is supervised on the query grid using the ground-truth mask resampled to the same resolution,
\begin{equation}
\mathcal L_{\rm pix}
=
\mathcal L_{\rm BCE}^{\rm pix}
+
\mathcal L_{\rm Dice}^{\rm pix}.
\label{eq:lpix}
\end{equation}
Both terms are computed separately for each target and then averaged across targets. The autoregressive response is trained with the language loss \(\mathcal L_{\rm text}\) over the generated target phrases and \texttt{[SEG]} tokens. The complete objective is
\begin{equation}
\mathcal L
=
\mathcal L_{\rm rep}
+
\mathcal L_{\rm pix}
+
\mathcal L_{\rm text}.
\label{eq:total-loss}
\end{equation}
All loss terms use unit coefficients and supervise language generation, mask-code prediction, sub-cell structure, and rendering.

At inference, greedy decoding generates the response, and the emitted \texttt{[SEG]} tokens determine the number of identified targets \(K\). The mask queries for all targets are evaluated in one mask pass with attention between different targets blocked. PACR renders a probability map for each target, which is resized to the original image resolution and thresholded at \(0.5\) to obtain the final binary mask. The discrete sub-cell mask \(\hat{\mathbf Y}_k\) can also be decoded from the same distribution over mask codes without continuous rendering. Additional implementation details are provided in the Supplement.

\section{Experiments}
\label{sec:experiments}

\subsection{Experimental Setup}

\paragraph{Implementation.}
We build \method{} on Qwen3-VL-4B \citep{bai2025qwen3vl}. Its vision connector merges each \(2\times2\) group of encoder patches into one visual token, with patch side length \(p=16\). Unless otherwise specified, we set the subdivision granularity to \(r=2\), the query-grid scale to \(\kappa=4\), and the channel width of PACR to \(256\). We use \(\method{}_{r\times r}\) to denote variants with \(r\in\{1,2,3\}\). The vision encoder is frozen during training. We optimize the language-model LoRA adapters,  the learnable parameters related to the registered special tokens, the mask-query projections, and PACR.

\paragraph{Benchmarks and metrics.}
We evaluate reasoning segmentation on LaSeRS \citep{xin2025segearth}, EarthReason \citep{li2025segearth}, and DRSeg \citep{ke2026pixdlm}. We evaluate explicit referring segmentation on RRSIS-D and RISBench \citep{liu2024rotated,dong2024cross}. Following prior work \citep{xin2025segearth}, we train benchmark-specific checkpoints and report gIoU and cIoU (\%). To examine the visual understanding retained after segmentation tuning, we compare the LaSeRS-trained checkpoint with its base MLLM on LHRS-Bench \citep{muhtar2024lhrs}. We measure mean inference time per sample and peak GPU memory on the LaSeRS test set using batch size \(1\) on a single NVIDIA A800 GPU. Additional implementation and training details are provided in the Supplement.

\subsection{Reasoning Segmentation in Remote Sensing Images}

\begin{figure*}[t]
\centering
\includegraphics[width=\textwidth]{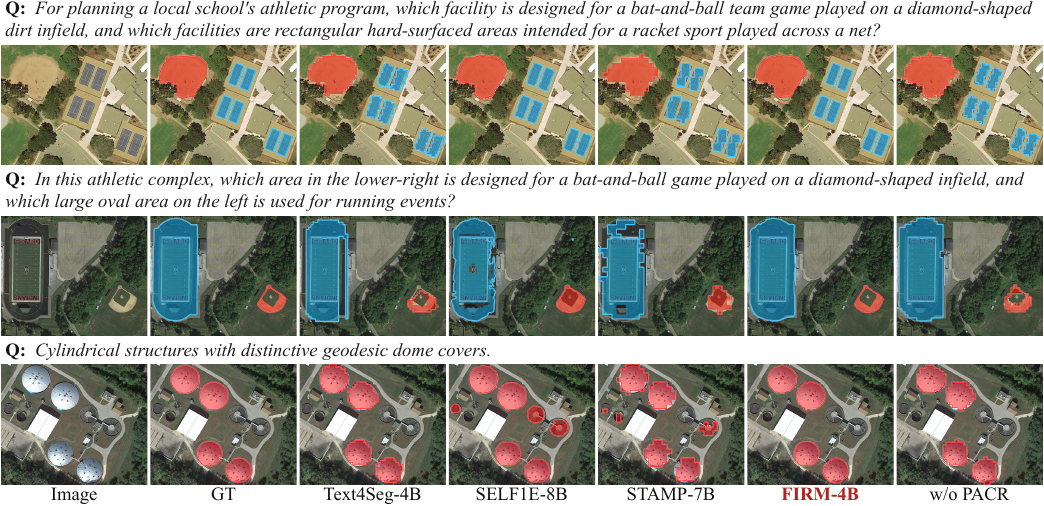}
\caption{Qualitative comparisons on remote sensing reasoning segmentation. The last column shows the discrete sub-cell mask produced by the same \method{} model before continuous rendering.}
\label{fig:qualitative}
\end{figure*}

LaSeRS evaluates reasoning segmentation across segmentation granularity, target multiplicity, reasoning requirement, and expression length. As shown in Table~\ref{tab:lasers}, \method{} achieves an average of \(70.5/80.5\) gIoU/cIoU and ranks first on 16 of the 18 subset metrics across methods with and without specialist mask decoders, although several competitors use larger backbones. Its largest margins occur for part-level targets, whose fine structures and boundaries are particularly sensitive to the spatial detail represented within individual visual tokens. This result is consistent with the motivation for explicitly representing intra-token mask patterns. STAMP provides the closest comparison because it also predicts a complete grid of binary mask tokens in one mask pass, but assigns only one binary label to each position. \method{} improves over STAMP-7B by \(20.7\) points in average gIoU, highlighting the benefit of representing a binary sub-cell pattern at each position. SELF1E-8B leads only on multiple-target and long-expression gIoU. Figure~\ref{fig:qualitative} further compares the predicted masks on thin structures and closely spaced targets.

\begin{table}[t]
\centering
\setlength{\tabcolsep}{2.0pt}
{\small
\begin{tabular}{@{}lccccc@{}}
\toprule
\multirow{2}{*}{Method} &
\multicolumn{2}{c}{Validation} & \multicolumn{2}{c}{Test} &
\multirow{2}{*}{Average} \\
\cmidrule(lr){2-3}\cmidrule(lr){4-5}
 & gIoU & cIoU & gIoU & cIoU & \\
\midrule
PixelLM & 57.9 & 57.8 & 60.0 & 59.2 & 58.7 \\
LISA & 61.0 & 57.4 & 60.9 & 59.1 & 59.6 \\
PSALM & 66.6 & 62.0 & 68.3 & 64.6 & 65.4 \\
Seg-Zero & 67.2 & 65.4 & 67.9 & 62.4 & 65.7 \\
VisionReasoner & 66.6 & 67.1 & 67.2 & 65.4 & 66.6 \\
SegEarth-R1 & 68.6 & 64.1 & 70.8 & 68.3 & 68.0 \\
RemoteReasoner & 69.0 & 67.8 & 71.0 & 69.1 & 69.2 \\
Text4Seg++ & 71.9 & 69.8 & 73.0 & 65.6 & 70.1 \\
SegEarth-R2-3B & \underline{72.3} & 68.1 & \underline{73.5} & 69.5 & 70.9 \\
Think2Seg-RS-3B & 69.2 & 71.3 & 70.8 & 74.2 & 71.4 \\
Think2Seg-RS-7B & 72.2 & \underline{72.9} & 73.4 & \underline{75.6} & \underline{73.5} \\
\midrule
\textbf{FIRM-4B} & \textbf{76.5} & \textbf{77.5} & \textbf{76.2} & \textbf{75.9} & \textbf{76.5} \\
\bottomrule
\end{tabular}
}
\caption{Comparison on the EarthReason dataset. Results are reported in \%; Average is the arithmetic mean of validation and test gIoU/cIoU.}
\label{tab:earthreason}
\end{table}

On EarthReason (Table~\ref{tab:earthreason}), where each method is trained separately, \method{} achieves the highest gIoU and cIoU on both validation and test splits, improving the average by \(3.0\) points over the strongest prior result. Its gains in both metrics demonstrate improvements in both per-sample mask overlap and pixel-aggregated overlap.
DRSeg evaluates reasoning segmentation in UAV images under attribute, scene, and spatial instructions. After supervised fine-tuning on DRSeg (Table~\ref{tab:drseg}), \method{} leads all six metrics and exceeds the previous SOTA method by more than \(11\) points on every subset and metric. These results indicate that the mask representation remains effective for UAV reasoning segmentation.

\begin{table}[t]
\centering
\setlength{\tabcolsep}{3.0pt}
{\small
\begin{tabular}{@{}lccc@{}}
\toprule
Method & Attribute & Scene & Spatial \\
\midrule
\multicolumn{4}{@{}l}{\textit{Zero-shot}} \\
SegEarth-R1 & 20.58/11.18 & 26.33/16.11 & 25.83/14.87 \\
LISA-13B & 52.65/48.12 & 47.08/42.41 & 42.85/37.94 \\
LISA-7B & 49.21/45.15 & 45.55/41.59 & 42.00/37.50 \\
LISA-7B (expl.) & 48.40/43.77 & 42.79/37.86 & 38.69/34.97 \\
LISAt-7B & 43.80/46.36 & 42.13/44.20 & 39.91/37.45 \\
PixelLM-7B & 46.87/49.55 & 43.07/45.49 & 41.28/40.94 \\
PixelLM-13B & 48.39/47.26 & 45.06/48.90 & 41.98/39.31 \\
\midrule
\multicolumn{4}{@{}l}{\textit{SFT}} \\
LISA-7B & 59.22/59.66 & 54.45/53.25 & 57.33/58.02 \\
PixelLM-7B & 58.39/57.35 & 55.19/54.29 & 55.38/54.47 \\
PixDLM & \underline{62.80}/\underline{62.84} & \underline{61.75}/\underline{64.03} & \underline{62.51}/\underline{62.80} \\
\textbf{FIRM-4B} & \textbf{76.79}/\textbf{77.44} & \textbf{74.13}/\textbf{75.19} & \textbf{74.98}/\textbf{75.15} \\
\bottomrule
\end{tabular}
}
\caption{Comparison on the DRSeg UAV test set under zero-shot transfer and benchmark-specific SFT. Each entry is gIoU/cIoU (\%).}
\label{tab:drseg}
\end{table}

\begin{table}[t]
\centering
\setlength{\tabcolsep}{0pt}
{\small
\begin{tabular*}{\columnwidth}{@{\extracolsep{\fill}}lccc@{}}
\toprule
Method & \shortstack{LaSeRS Avg.\\gIoU/cIoU \(\uparrow\)} &
\shortstack{Time (s)\\\(\downarrow\)} &
\shortstack{Peak VRAM\\(GB) \(\downarrow\)} \\
\midrule
\multicolumn{4}{@{}l}{\textit{Published native outputs}} \\
Text4Seg-4B & 52.6/56.2 & 54.26 & 10.57 \\
UFO-8B & 50.6/61.2 & 2.50 & 17.79 \\
SELF1E-2B & 63.3/73.1 & 1.71 & 4.98 \\
SELF1E-8B & 68.7/77.1 & 1.92 & 17.76 \\
STAMP-2B & 46.7/64.2 & 0.99 & 4.91 \\
STAMP-7B & 49.8/65.7 & 1.97 & 19.87 \\
\midrule
\multicolumn{4}{@{}l}{\textit{\method{} variants}} \\
\(\method{}_{1\times1}\) & 62.8/76.1 & 1.42 & 9.95 \\
\(\method{}_{1\times1}\) w/o PACR & 46.2/66.2 & 1.38 & 9.91 \\
\(\method{}_{2\times2}\) & 70.5/80.5 & 1.43 & 10.35 \\
\(\method{}_{2\times2}\) w/o PACR & 59.1/73.7 & 1.40 & 9.91 \\
\(\method{}_{3\times3}\) & 76.1/83.0 & 1.46 & 12.01 \\
\(\method{}_{3\times3}\) w/o PACR & 64.1/76.6 & 1.41 & 9.91 \\
\bottomrule
\end{tabular*}
}
\caption{Ablation of intra-token subdivision granularity and PACR on LaSeRS. We report average gIoU/cIoU (\%), mean inference time per sample, and peak GPU memory.}
\label{tab:interfaces}
\end{table}

\begin{table}[!t]
\centering
\setlength{\tabcolsep}{2.5pt}
{\small
\begin{tabular}{@{}l*{4}{r@{/}l}@{}}
\toprule
\multirow{2}{*}{Method} & \multicolumn{4}{c}{RRSIS-D} & \multicolumn{4}{c}{RISBench} \\
\cmidrule(lr){2-5}\cmidrule(lr){6-9}
 & \multicolumn{2}{c}{Validation} & \multicolumn{2}{c}{Test} &
 \multicolumn{2}{c}{Validation} & \multicolumn{2}{c}{Test} \\
\midrule
\multicolumn{9}{@{}l}{\textit{Segmentation specialists}} \\
RIS-DMMI & 60.7 & 77.0 & 60.1 & 76.2 & 62.6 & \underline{70.6} & 63.9 & \underline{74.8} \\
LAVT & 61.5 & 77.6 & 61.0 & 77.2 & 60.5 & 69.4 & 61.9 & 74.2 \\
RMSIN & 65.1 & 78.3 & 64.2 & 77.8 & 61.8 & 69.5 & 63.1 & 74.1 \\
\midrule
\multicolumn{9}{@{}l}{\textit{MLLM-based methods}} \\
LISA & 27.8 & \(-\) & 26.8 & \(-\) & \(-\) & \(-\) & \(-\) & \(-\) \\
PixelLM & 33.9 & \(-\) & 31.7 & \(-\) & \(-\) & \(-\) & \(-\) & \(-\) \\
GeoGround & 61.1 & \(-\) & 60.5 & \(-\) & \(-\) & \(-\) & \(-\) & \(-\) \\
GeoPixel & 68.0 & \textbf{81.8} & 67.3 & \textbf{84.9} & \(-\) & \(-\) & \(-\) & \(-\) \\
SegEarth-R1 & 67.6 & 78.9 & 66.4 & 78.0 & \(-\) & \(-\) & \(-\) & \(-\) \\
Text4Seg++ & 64.1 & 75.8 & 62.8 & 74.2 & \(-\) & \(-\) & \(-\) & \(-\) \\
SegEarth-R2 & \underline{68.8} & \(-\) & \underline{67.9} & \(-\) & \underline{69.8} & \(-\) & \underline{70.5} & \(-\) \\
\midrule
\textbf{FIRM-4B} & \textbf{70.8} & \underline{81.4} & \textbf{70.3} & \underline{81.8} & \textbf{70.2} & \textbf{73.8} & \textbf{72.1} & \textbf{78.8} \\
\bottomrule
\end{tabular}
}
\caption{Comparison on explicit remote sensing referring segmentation. Each entry is gIoU/cIoU (\%); a dash marks a metric the method does not report.}
\label{tab:referring}
\end{table}

\subsection{Representation Granularity and Continuous Rendering}

Table~\ref{tab:interfaces} evaluates subdivision granularity and continuous rendering while keeping the backbone, training recipe, and number of mask queries fixed. Increasing \(r\) consistently improves both the discrete and rendered masks. For the discrete readout, increasing \(r\) from \(1\) to \(2\) improves gIoU by \(12.9\) points, followed by another \(5.0\)-point gain at \(r=3\). The \(r=3\) sub-cell grid is finer than the encoder patch grid, showing that the output subdivision granularity need not be tied to the connector merge factor.
PACR provides gains of \(16.6\), \(11.4\), and \(12.0\) gIoU at \(r=1,2,3\), respectively. Its largest improvement occurs when the discrete grid is coarsest, while substantial gains remain at finer granularities. Moreover, the discrete result at each of \(r=2\) and \(r=3\) remains below the rendered result at the preceding granularity. These results show that a finer intra-token representation and continuous boundary refinement contribute complementary improvements.
The six \method{} variants differ by at most \(0.08\) seconds in inference time. At the default \(r=2\), PACR adds \(0.03\) seconds and \(0.44\) GB of peak memory. In the measured setting, \(\method{}_{2\times2}\) is faster and uses less memory than SELF1E-8B and STAMP-7B while achieving higher accuracy. Text4Seg-4B requires substantially more inference time, consistent with its autoregressive generation of textual mask labels.

\subsection{Referring Segmentation in Remote Sensing Images}

Referring segmentation provides a complementary setting in which the instruction identifies the target explicitly. As shown in Table~\ref{tab:referring}, \method{} achieves the highest results on all four RISBench metrics. On RRSIS-D, it achieves the highest validation and test gIoU, while ranking second in cIoU behind GeoPixel, an 8B model equipped with a specialist mask decoder. The stronger gIoU indicates more consistent per-sample mask overlap, whereas GeoPixel retains higher overlap when intersections and unions are accumulated across the dataset. These results demonstrate that the intra-token mask representation also applies effectively to explicit instructions.

\subsection{Visual Understanding Retention}

\begin{table}[t]
\centering
\setlength{\tabcolsep}{0pt}
{\small
\begin{tabular*}{\columnwidth}{@{\extracolsep{\fill}}llccc@{}}
\toprule
\multirow{2}{*}{Method} & \multirow{2}{*}{LLM} &
\multicolumn{2}{c}{LHRS-Bench} &
LaSeRS Avg. \\
\cmidrule(lr){3-4}
 & & Base & Tuned & gIoU/cIoU \\
\midrule
STAMP-7B & Qwen2-VL-7B & 69.06 & 38.71 & 49.8/65.7 \\
STAMP-2B & Qwen2-VL-2B & 54.18 & 25.98 & 46.7/64.2 \\
SegEarth-R2-3B & Mipha-3B & 65.43 & 49.59 & 57.2/67.9 \\
Text4Seg-4B & Qwen3-VL-4B & 74.54 & 61.51 & 52.6/56.2 \\
UFO-8B & InternVL2.5-8B & 67.14 & 45.89 & 50.6/61.2 \\
SELF1E-2B & InternVL3-2B & 66.40 & 38.56 & 63.3/73.1 \\
SELF1E-8B & InternVL3-8B & 69.28 & 47.89 & 68.7/77.1 \\
\midrule
\textbf{FIRM-4B} & Qwen3-VL-4B & 74.54 & 68.25 & 70.5/80.5 \\
\bottomrule
\end{tabular*}
}
\caption{LHRS-Bench micro-average accuracy (\%) before and after LaSeRS segmentation tuning, together with the average LaSeRS gIoU/cIoU (\%) of each tuned checkpoint.}
\label{tab:lhrs}
\end{table}

Table~\ref{tab:lhrs} evaluates how much general remote sensing visual understanding is retained after segmentation tuning. Every tuned checkpoint performs worse than its base MLLM on LHRS-Bench, with accuracy decreases ranging from \(30.4\) points for STAMP-7B to \(6.3\) points for \method{}. \method{} retains an accuracy of \(68.25\) from its \(74.54\) base model while also achieving the highest LaSeRS result.
Text4Seg provides the most informative comparison because it uses the same Qwen3-VL-4B model. Its tuned checkpoint reaches \(61.51\) on LHRS-Bench and \(52.6/56.2\) gIoU/cIoU on LaSeRS, compared with \(68.25\) and \(70.5/80.5\) for \method{}. This pattern is consistent with mask-code prediction through the language-model output interface preserving more of the original visual understanding while learning fine-grained segmentation.

\section{Conclusion}
\label{sec:conclusion}

Through \method{}, this work examines how mask representation interacts with visual-token compression in MLLM-based segmentation. Results across five benchmarks show that preserving spatial variation within each compressed visual token substantially improves mask quality in both reasoning and referring settings. The granularity study further reveals consistent gains as the intra-token representation becomes finer. Continuous rendering provides additional improvements at every evaluated granularity, showing that its benefit remains even with a finer discrete representation. Together, these findings establish mask representation as an important complement to semantic reasoning. Identifying the correct target is only the first step, and the model must also retain sufficient spatial expressiveness to delineate it precisely. This perspective points toward richer spatial output representations for future MLLMs.

\bibliography{aaai2027}

\appendix
\twocolumn[
\begin{center}
{\Large\bfseries Supplementary Material for
\method{}: Fine-Grained Intra-Token Representation of Masks for Remote Sensing Reasoning Segmentation\par}
\end{center}
]

\section{Expected Hamming Loss}

Expected Hamming supervision transfers the geometry of binary sub-cell patterns to the categorical mask-code space. For a ground-truth code \(c^\star_{k,j}\), the Hamming distance of a candidate code \(c\) is
\begin{equation}
d_H(c,c^\star_{k,j})
=
\sum_{u=1}^{r^2}
\left|b_u(c)-b_u(c^\star_{k,j})\right|.
\label{eq:app-ham-distance}
\end{equation}
Let \(y_{k,j,u}=b_u(c^\star_{k,j})\), and denote the unweighted inner expectation in Equation~(\ref{eq:lham}) at position \((k,j)\) by \(\ell_{\rm Ham}^{k,j}\). Since \(y_{k,j,u}\) is binary, the expected disagreement at sub-cell \(u\) is \(y_{k,j,u}(1-\mu_{k,j,u})+(1-y_{k,j,u})\mu_{k,j,u}=|\mu_{k,j,u}-y_{k,j,u}|\). Exchanging the sums over codes and sub-cells then gives
\begin{equation}
\ell_{\rm Ham}^{k,j}
=
\sum_{u=1}^{r^2}
\left|\mu_{k,j,u}-y_{k,j,u}\right|.
\label{eq:app-ham-marginal}
\end{equation}
The identity gives expected Hamming loss a direct structural interpretation: it is the sum of marginal foreground errors within a mask code. It retains prediction over complete code patterns while assigning smaller penalties to alternatives that differ in fewer sub-cells.

\section{Additional Training Settings}

Table~\ref{tab:app-final-config} summarizes the additional settings used in training.

\noindent\begin{minipage}{\columnwidth}
\centering
\setlength{\tabcolsep}{3pt}
{\small
\begin{tabular}{@{}>{\raggedright\arraybackslash}p{2.0cm}>{\raggedright\arraybackslash}p{5.45cm}@{}}
\toprule
Component & Setting \\
\midrule
Input and adaptation &
Input pixel budget \(2^{18}\)--\(2^{20}\); bfloat16 training; rank-64 LoRA with \(\alpha=128\) and dropout \(0.05\). \\
Optimization &
AdamW for \(5\) epochs with batch size \(64\), learning rate \(2\times10^{-4}\), cosine decay to \(2\times10^{-6}\), \(3\%\) learning-rate warmup, \((\beta_1,\beta_2)=(0.9,0.95)\), zero weight decay, and gradient clipping at \(1.0\). \\
Hardware &
\(4\times\) NVIDIA A800 GPUs (\(80\,\mathrm{GB}\)). \\
Random seed &
\(42\). \\
Supervision &
Sub-cell occupancy threshold \(\tau=0.5\), background-code weight \(0.1\), other code weights \(1.0\), and Dice smoothing \(\epsilon=1\). \\
\bottomrule
\end{tabular}
}
\captionof{table}{Additional settings used in training.}
\label{tab:app-final-config}
\end{minipage}

\paragraph{LaSeRS baseline reproduction.}
The UFO~\citep{tang2025ufo}, SELF1E~\citep{zhang2026self1e}, STAMP~\citep{liu2026stamp}, and Text4Seg~\citep{lan2024text4seg} entries in Table~\ref{tab:lasers} are our LaSeRS reproductions based on the official implementations released by their authors. We train and evaluate all four methods on LaSeRS under the protocol used in Table~\ref{tab:lasers}. Text4Seg does not prescribe a fixed MLLM backbone; we instantiate it with Qwen3-VL-4B, represent each mask on a high-resolution \(64\times64\) semantic-descriptor grid, and report the resulting model as Text4Seg-4B.

\section{Prompt Construction and Target Alignment}

Prompt construction follows three families. LaSeRS retains its dataset-provided multi-target response. EarthReason and DRSeg share a reasoning-segmentation user turn but use different answer forms. RRSIS-D and RISBench share one referring-expression family, with the user and assistant turns sampled independently from six question-template entries and ten answer-template entries during training. Tables~\ref{tab:app-prompt-lasers}--\ref{tab:app-answer-templates} show the prompt formats and the complete referring-expression template pools.

Mask generation is triggered by \texttt{[SEG]} tokens in the assistant response. LaSeRS may emit multiple triggers, with the \(k\)-th target phrase--\texttt{[SEG]} pair aligned with the \(k\)-th annotated mask. The other four datasets contain one target per sample and therefore use one response trigger.

\begingroup
\definecolor{PromptFrame}{HTML}{4A4A4A}
\definecolor{PromptBg}{HTML}{F3F4F6}
\definecolor{PromptUser}{HTML}{006D77}
\definecolor{PromptAssistant}{HTML}{7A3E00}
\definecolor{PromptImage}{HTML}{005FAD}
\definecolor{PromptTarget}{HTML}{1B6E3A}
\definecolor{PromptAnswer}{HTML}{8A3B76}
\definecolor{PromptSeg}{HTML}{B34400}
\setlength{\fboxrule}{0.8pt}
\setlength{\fboxsep}{6pt}

\noindent\begin{minipage}{\columnwidth}
\centering
\fcolorbox{PromptFrame}{PromptBg}{\begin{minipage}{0.91\columnwidth}
\raggedright
\small
{\color{PromptUser}\bfseries\sffamily USER}\par
\smallskip
{\color{PromptImage}\texttt{\detokenize{<image>}}}\par
{\color{PromptTarget}\texttt{\detokenize{<raw LaSeRS question>}}}

\medskip
{\color{PromptAssistant}\bfseries\sffamily ASSISTANT}\par
\smallskip
Natural-language response containing
{\color{PromptTarget}\texttt{\detokenize{<p> target_k </p>}}}
{\color{PromptSeg}\texttt{[SEG]}\(_k\)}, for \(k=1,\ldots,K\).
\end{minipage}}
\captionof{table}{LaSeRS multi-target prompt.}
\label{tab:app-prompt-lasers}
\end{minipage}

\medskip

\noindent\begin{minipage}{\columnwidth}
\centering
\fcolorbox{PromptFrame}{PromptBg}{\begin{minipage}{0.91\columnwidth}
\raggedright
\small
{\color{PromptUser}\bfseries\sffamily USER: SHARED}\par
\smallskip
This is an image {\color{PromptImage}\texttt{\detokenize{<image>}}}, Please doing
Reasoning Segmentation according to the following instruction:\par
{\color{PromptTarget}\texttt{\detokenize{<raw reasoning question>}}}

\medskip
{\color{PromptAssistant}\bfseries\sffamily ASSISTANT: EARTHREASON}\par
\smallskip
{\color{PromptAnswer}\texttt{\detokenize{<description answer>}}} The corresponding region is
{\color{PromptSeg}\texttt{[SEG]}}.

\medskip
{\color{PromptAssistant}\bfseries\sffamily ASSISTANT: DRSEG}\par
\smallskip
{\color{PromptAnswer}\texttt{\detokenize{<short reasoning answer>}}}
{\color{PromptSeg}\texttt{[SEG]}}
\end{minipage}}
\captionof{table}{Single-target reasoning prompts for EarthReason and DRSeg. The user turn is shared, while the assistant response is dataset specific.}
\label{tab:app-prompt-reasoning}
\end{minipage}

\medskip

\noindent\begin{minipage}{\columnwidth}
\centering
\fcolorbox{PromptFrame}{PromptBg}{\begin{minipage}{0.91\columnwidth}
\raggedright
\small
{\color{PromptUser}\bfseries\sffamily USER}\par
\smallskip
{\color{PromptImage}\texttt{\detokenize{<image>}}}\par
Please segment only the
{\color{PromptTarget}\texttt{\detokenize{<referring expression>}}} in the image.

\medskip
{\color{PromptAssistant}\bfseries\sffamily ASSISTANT}\par
\smallskip
Sure, here is the segmentation mask for
'{\color{PromptTarget}\texttt{\detokenize{<referring expression>}}}':
{\color{PromptSeg}\texttt{[SEG]}}
\end{minipage}}
\captionof{table}{Shared referring-expression prompt family for RRSIS-D and RISBench.}
\label{tab:app-prompt-referring}
\end{minipage}

\medskip

\newcommand{\appPromptSlot}{{\color{PromptTarget}\texttt{\detokenize{[class_name]}}}}

\noindent\begin{minipage}{\columnwidth}
\centering
\fcolorbox{PromptFrame}{PromptBg}{\begin{minipage}{0.91\columnwidth}
\raggedright
\small
{\color{PromptUser}\bfseries\sffamily QUESTION TEMPLATES}\par
\smallskip
{\color{PromptUser}\bfseries Q1}\quad Please segment only the \appPromptSlot{} in the image.\par\smallskip
{\color{PromptUser}\bfseries Q2}\quad Can you segment the \appPromptSlot{} in the image?\par\smallskip
{\color{PromptUser}\bfseries Q3}\quad Where is the \appPromptSlot{} in this picture? Please respond with segmentation mask.\par\smallskip
{\color{PromptUser}\bfseries Q4}\quad Where is '\appPromptSlot{}' in this image? Please output segmentation mask.\par\smallskip
{\color{PromptUser}\bfseries Q5}\quad Could you provide the segmentation mask for '\appPromptSlot{}' in this image?\par\smallskip
{\color{PromptUser}\bfseries Q6}\quad Please segment the image and highlight '\appPromptSlot{}'.
\end{minipage}}
\captionof{table}{Complete question-template pool for RRSIS-D and RISBench. The placeholder is replaced by the referring expression in each sample.}
\label{tab:app-question-templates}
\end{minipage}

\medskip

\noindent\begin{minipage}{\columnwidth}
\centering
\fcolorbox{PromptFrame}{PromptBg}{\begin{minipage}{0.91\columnwidth}
\raggedright
\small
{\color{PromptAssistant}\bfseries\sffamily ANSWER TEMPLATE ENTRIES}\par
\smallskip
{\color{PromptAssistant}\bfseries A1}\quad Sure, here is the segmentation mask for '\appPromptSlot{}':\par\smallskip
{\color{PromptAssistant}\bfseries A2}\quad Here is the segmentation map focusing on the \appPromptSlot{}:\par\smallskip
{\color{PromptAssistant}\bfseries A3}\quad Here is the segmentation mask highlighting the \appPromptSlot{}:\par\smallskip
{\color{PromptAssistant}\bfseries A4}\quad The segmentation map for '\appPromptSlot{}' is:\par\smallskip
{\color{PromptAssistant}\bfseries A5}\quad The segmentation mask for '\appPromptSlot{}' is shown below:\par\smallskip
{\color{PromptAssistant}\bfseries A6}\quad Sure, Here's the segmentation of the \appPromptSlot{}:\par\smallskip
{\color{PromptAssistant}\bfseries A7}\quad Sure, the segmented output for '\appPromptSlot{}' is:\par\smallskip
{\color{PromptAssistant}\bfseries A8}\quad Certainly, the segmentation map for '\appPromptSlot{}' is:\par\smallskip
{\color{PromptAssistant}\bfseries A9}\quad Certainly, here is the segmentation mask for '\appPromptSlot{}':\par\smallskip
{\color{PromptAssistant}\bfseries A10}\quad The segmentation mask for '\appPromptSlot{}' is shown below:\par\smallskip
{\color{PromptSeg}\texttt{[SEG]}} is appended to every sampled answer.
\end{minipage}}
\captionof{table}{Complete ten-entry answer-template pool for RRSIS-D and RISBench.}
\label{tab:app-answer-templates}
\end{minipage}
\endgroup

\section{Evaluation Metrics and Multi-Target Protocol}

Let \(A_t\) and \(B_t\) be the predicted and ground-truth binary masks of scored unit \(t\), with ground-truth pixels assigned the ignore label \(255\) excluded. Write \(I_t=|A_t\cap B_t|\) and \(U_t=|A_t\cup B_t|\). The IoU of a scored unit is
\begin{equation}
{\rm IoU}_t
=
\frac{I_t}{U_t},
\qquad U_t>0.
\label{eq:app-iou}
\end{equation}
When \(U_t=0\), the prediction and target are both empty and we set \({\rm IoU}_t=1\).
The reported gIoU is the mean across the \(T\) scored units,
\begin{equation}
{\rm gIoU}
=
\frac{1}{T}
\sum_{t=1}^{T}
{\rm IoU}_t.
\label{eq:app-giou}
\end{equation}
The reported cIoU accumulates intersections and unions before division,
\begin{equation}
{\rm cIoU}
=
\frac{
\sum_{t=1}^{T} I_t
}{
\sum_{t=1}^{T} U_t + \epsilon_{\rm IoU}
}.
\label{eq:app-ciou}
\end{equation}
We use \(\epsilon_{\rm IoU}=10^{-5}\). gIoU gives equal weight to each scored unit, while cIoU weights units according to their union area and therefore emphasizes pixel-aggregated overlap.

For LaSeRS multi-target samples, we follow the ordered protocol of SegEarth-R2~\citep{xin2025segearth}: predictions are matched to annotations in generation order, with the last prediction repeated when fewer predictions than annotations are generated and excess predictions truncated. Results in Table~\ref{tab:lasers} are computed using this protocol. Its \emph{Avg.} column is the unweighted macro average of the nine official LaSeRS subsets, computed after obtaining gIoU and cIoU within each subset.

\section{Loss Ablation}

Table~\ref{tab:app-loss-ablation} ablates the representation loss \(\mathcal L_{\rm rep}\) and pixel loss \(\mathcal L_{\rm pix}\) defined in Equations~(\ref{eq:lrep}) and~(\ref{eq:lpix}). The complete objective achieves the highest LaSeRS average, and removing any individual loss term reduces both metrics. Removing \(\mathcal L_{\rm Dice}^{\rm pix}\) produces the largest individual decrease, lowering gIoU and cIoU by 4.50 and 2.04 points, respectively. Removing all three terms in \(\mathcal L_{\rm rep}\) lowers gIoU and cIoU by 2.99 and 3.73 points, showing that mask-code supervision complements pixel-level supervision of the rendered mask.

\begin{table}[t]
\centering
\setlength{\tabcolsep}{2pt}
{\small
\begin{tabular*}{\columnwidth}{@{\extracolsep{\fill}}cccccc@{}}
\toprule
\multicolumn{3}{c}{\(\mathcal L_{\rm rep}\)} &
\multicolumn{2}{c}{\(\mathcal L_{\rm pix}\)} &
LaSeRS Avg. \\
\cmidrule(lr){1-3}\cmidrule(lr){4-5}
\(\mathcal L_{\rm CE}\) &
\(\mathcal L_{\rm Ham}\) &
\(\mathcal L_{\rm Dice}^{\rm str}\) &
\(\mathcal L_{\rm BCE}^{\rm pix}\) &
\(\mathcal L_{\rm Dice}^{\rm pix}\) &
gIoU/cIoU \\
\midrule
\(\times\) & \(\checkmark\) & \(\checkmark\) & \(\checkmark\) & \(\checkmark\) & 68.59 / 78.97 \\
\(\checkmark\) & \(\times\) & \(\checkmark\) & \(\checkmark\) & \(\checkmark\) & 69.28 / 79.24 \\
\(\checkmark\) & \(\checkmark\) & \(\times\) & \(\checkmark\) & \(\checkmark\) & 69.70 / 80.10 \\
\(\checkmark\) & \(\checkmark\) & \(\checkmark\) & \(\times\) & \(\checkmark\) & 69.94 / 80.05 \\
\(\checkmark\) & \(\checkmark\) & \(\checkmark\) & \(\checkmark\) & \(\times\) & 66.00 / 78.46 \\
\(\times\) & \(\times\) & \(\times\) & \(\checkmark\) & \(\checkmark\) & 67.51 / 76.77 \\
\midrule
\(\checkmark\) & \(\checkmark\) & \(\checkmark\) & \(\checkmark\) & \(\checkmark\) & \textbf{70.50 / 80.50} \\
\bottomrule
\end{tabular*}
}
\caption{Ablation study of the loss components on LaSeRS. Results are reported in \%.}
\label{tab:app-loss-ablation}
\end{table}

\section{Additional Qualitative Analysis}

\begin{figure*}[t]
\centering
\includegraphics[width=\textwidth]{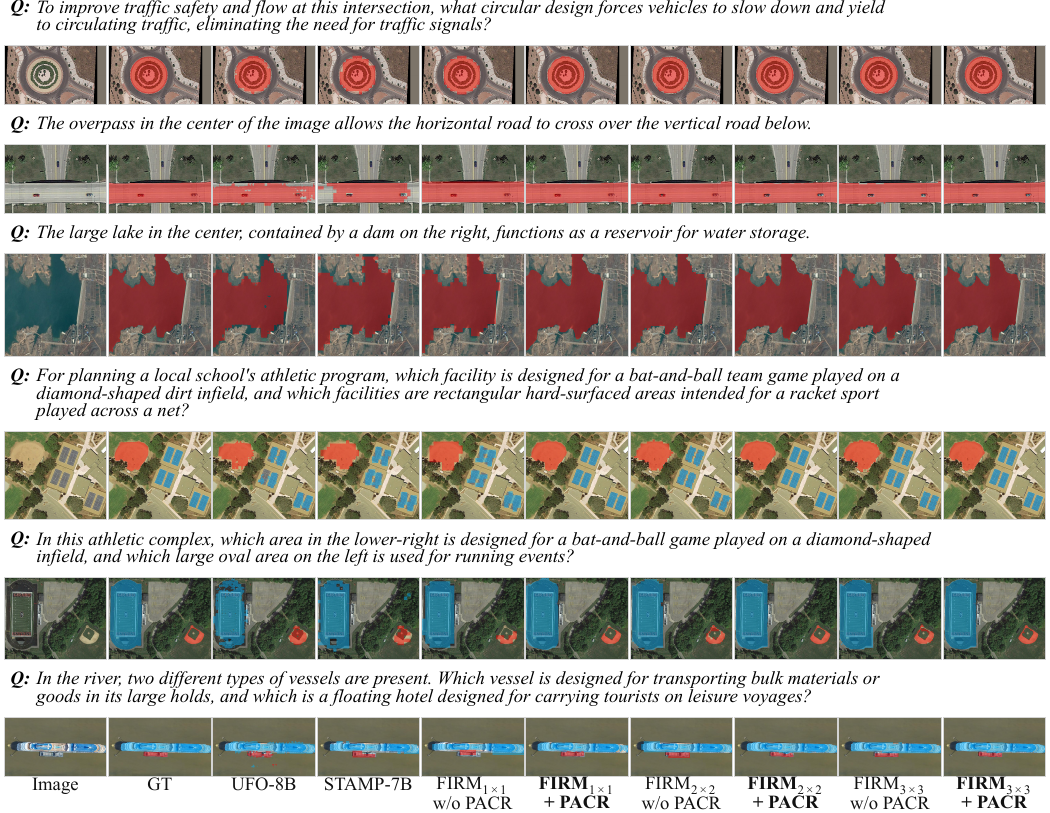}
\caption{Additional qualitative comparisons. Each row shows the input image, ground-truth mask, UFO-8B, STAMP-7B, and \method{} at \(r=1,2,3\), with and without PACR. The comparisons illustrate the effects of subdivision granularity and continuous rendering on the predicted masks.}
\label{fig:app-qualitative-comparison}
\end{figure*}

Figure~\ref{fig:app-qualitative-comparison} compares the reproduced UFO-8B and STAMP-7B baselines with \method{} variants at \(r=1,2,3\), with and without PACR. These examples complement Table~\ref{tab:interfaces}: increasing \(r\) yields finer discrete sub-cell masks, while PACR refines the soft structural field into rendered masks with more continuous boundaries.

\section{Qualitative Question Answering and Grounding}

\begin{figure*}[p]
\centering
\includegraphics[width=\textwidth]{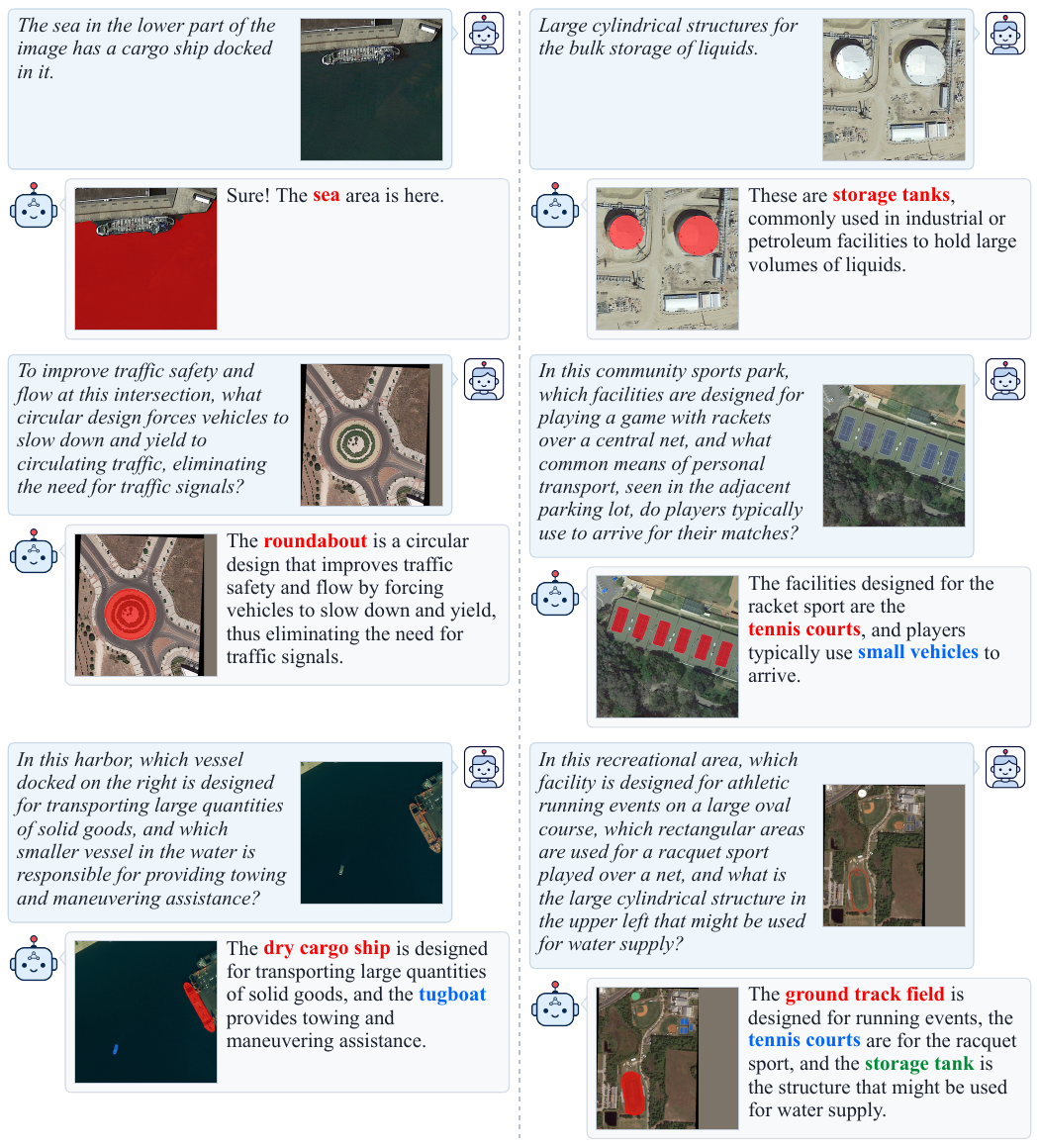}
\caption{Qualitative results of \method{} on LaSeRS.}
\label{fig:app-qa-lasers}
\end{figure*}

\begin{figure*}[p]
\centering
\includegraphics[width=\textwidth]{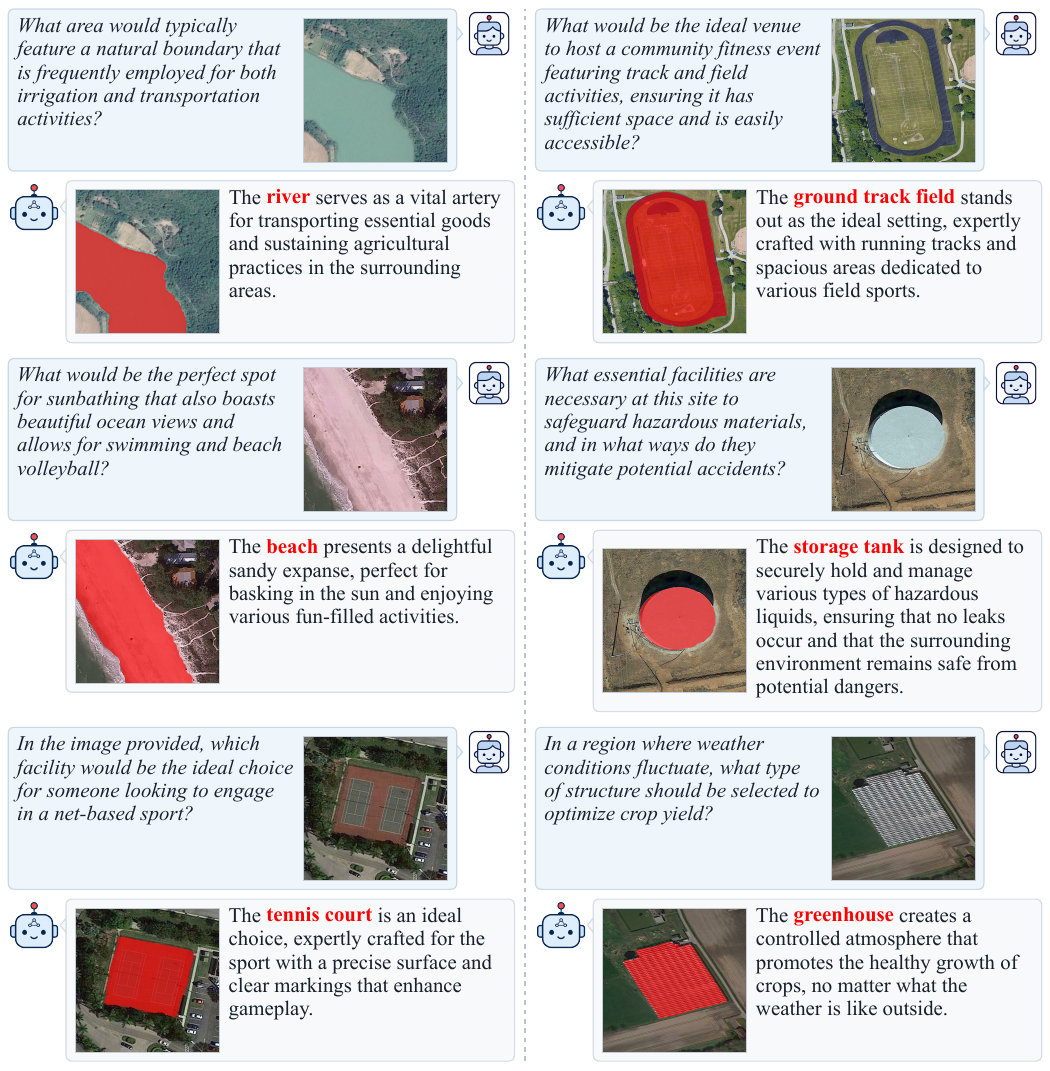}
\caption{Qualitative results of \method{} on EarthReason.}
\label{fig:app-qa-earthreason}
\end{figure*}

\begin{figure*}[p]
\centering
\includegraphics[width=\textwidth]{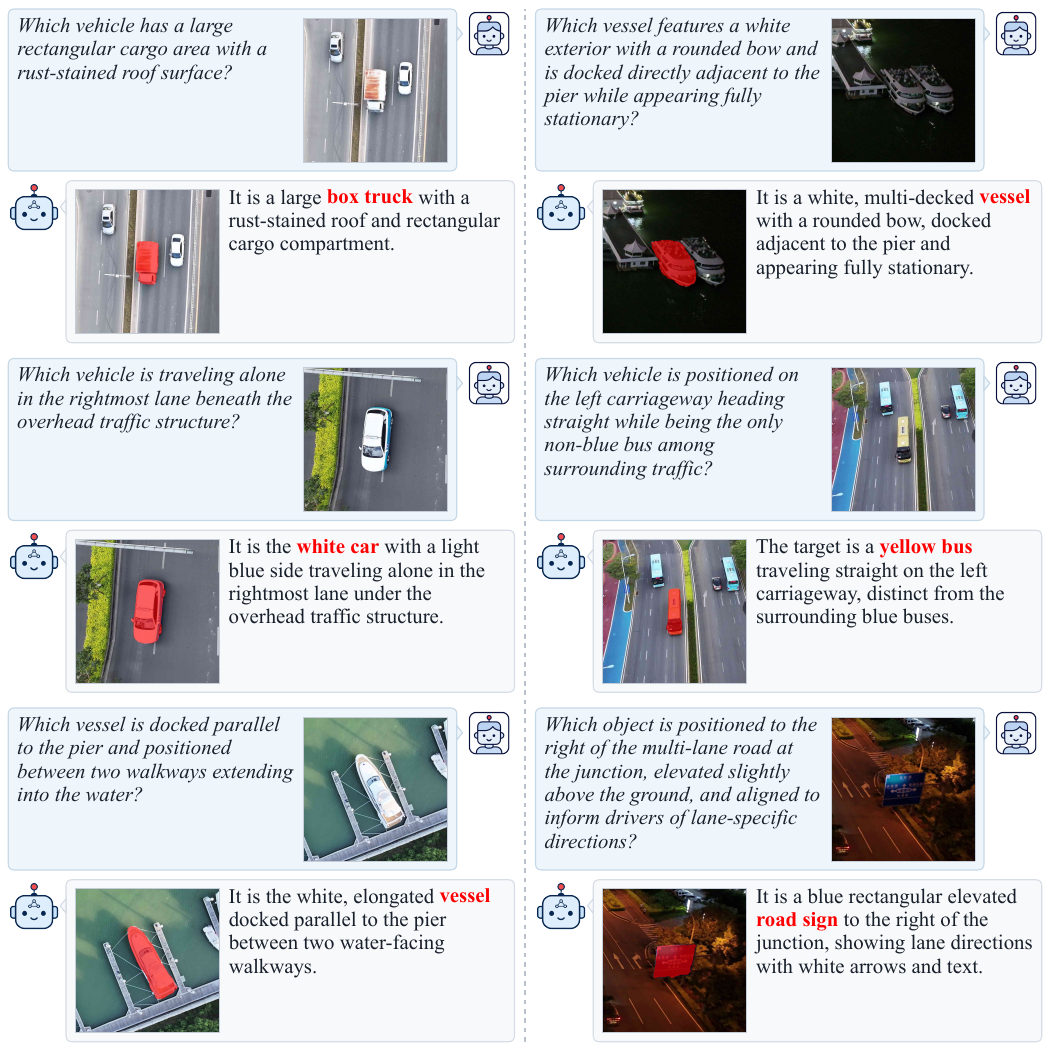}
\caption{Qualitative results of \method{} on DRSeg.}
\label{fig:app-qa-drseg}
\end{figure*}

\begin{figure*}[p]
\centering
\includegraphics[width=\textwidth]{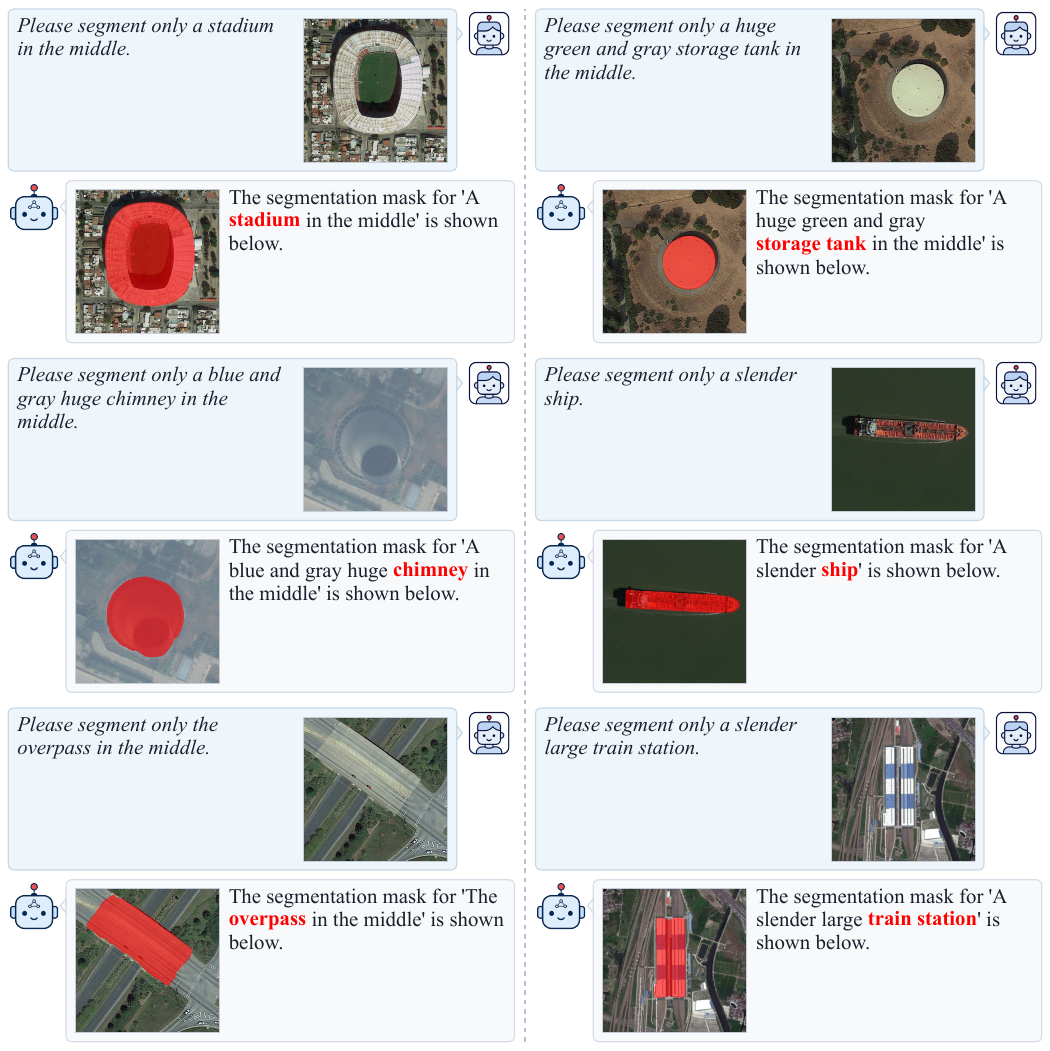}
\caption{Qualitative results of \method{} on RRSIS-D.}
\label{fig:app-qa-rrsisd}
\end{figure*}

\begin{figure*}[p]
\centering
\includegraphics[width=\textwidth]{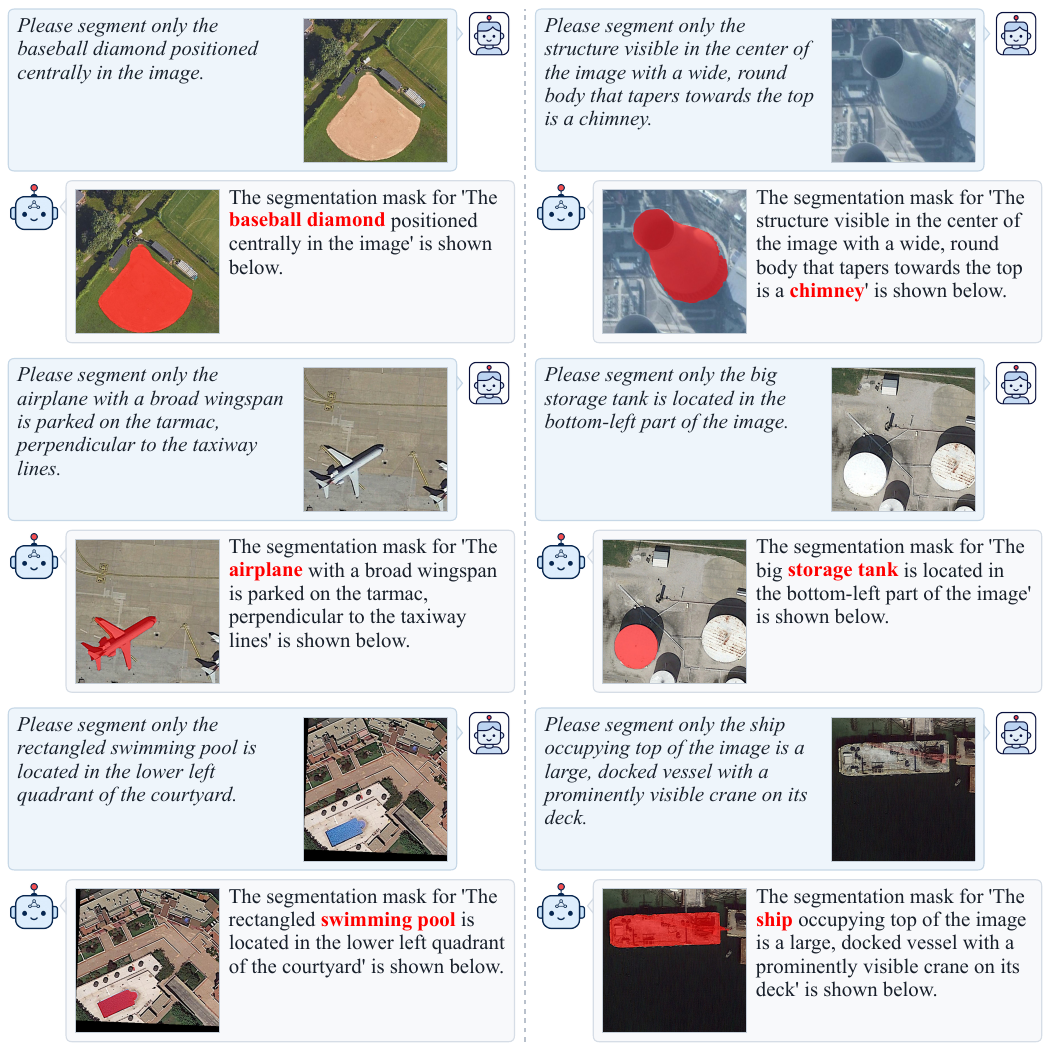}
\caption{Qualitative results of \method{} on RISBench.}
\label{fig:app-qa-risbench}
\end{figure*}

Figures~\ref{fig:app-qa-lasers}--\ref{fig:app-qa-risbench} present qualitative results of \method{} on LaSeRS, EarthReason, DRSeg, RRSIS-D, and RISBench, respectively. Each figure shows the generated responses together with their corresponding predicted masks.

\section{Multi-Turn Referring Segmentation Examples}

\begin{figure*}[p]
\centering
\includegraphics[width=\textwidth]{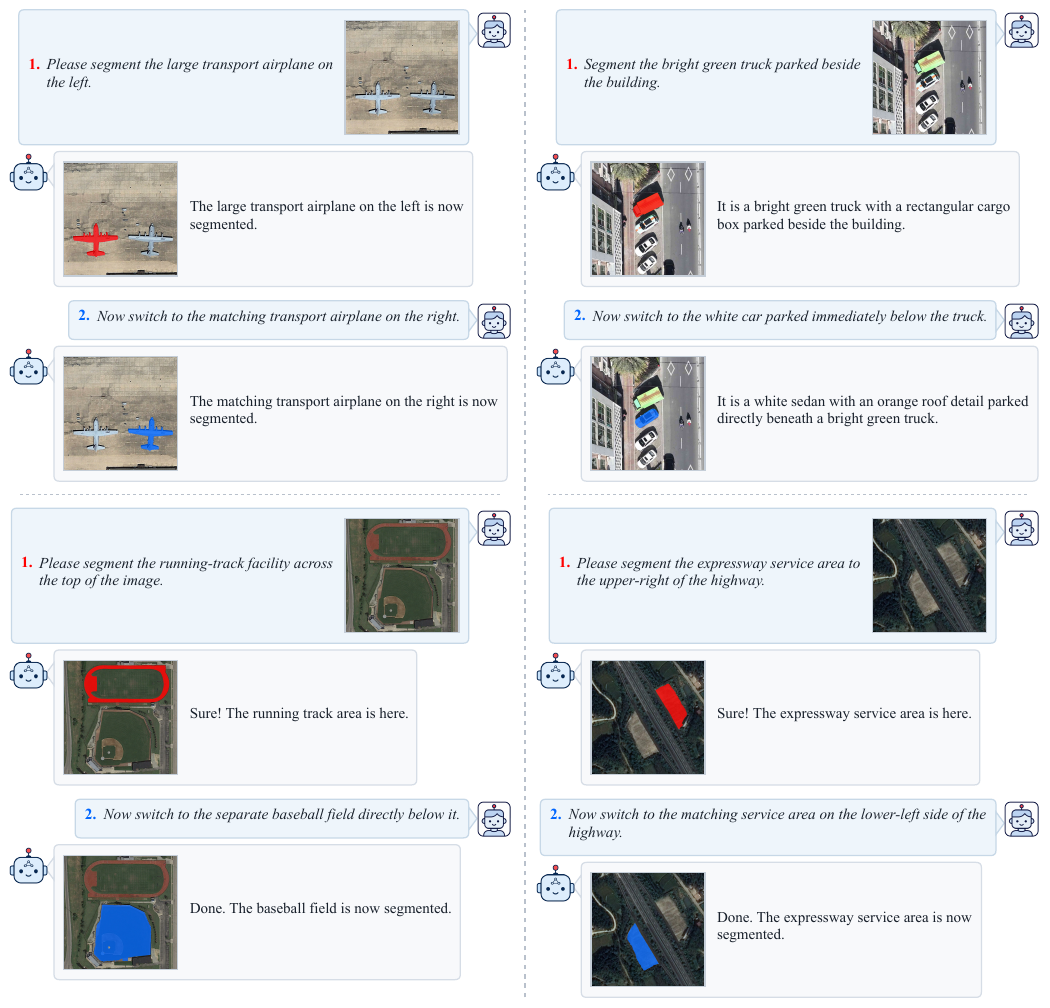}
\caption{Qualitative multi-turn referring segmentation examples (Part I).}
\label{fig:app-multiturn-1}
\end{figure*}

\begin{figure*}[p]
\centering
\includegraphics[width=\textwidth]{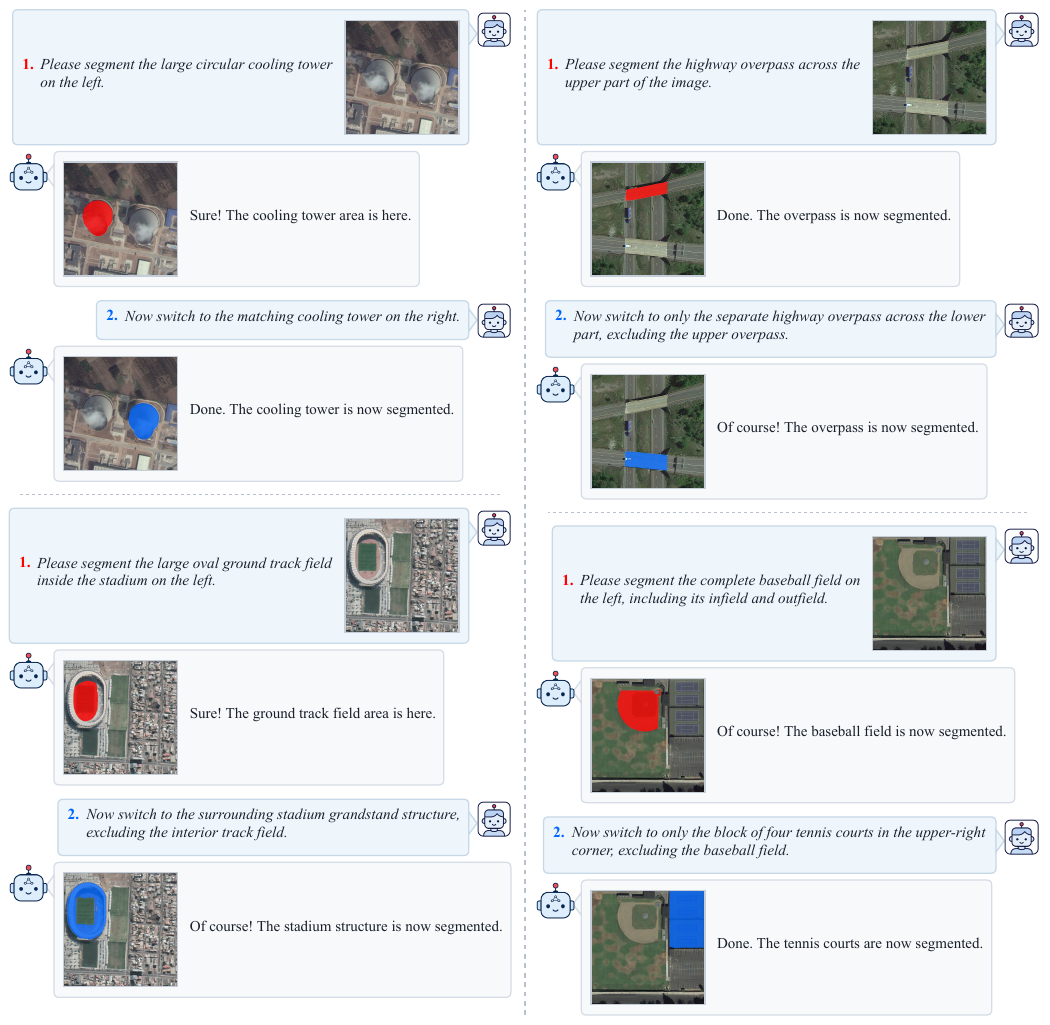}
\caption{Qualitative multi-turn referring segmentation examples (Part II).}
\label{fig:app-multiturn-2}
\end{figure*}

Figures~\ref{fig:app-multiturn-1} and~\ref{fig:app-multiturn-2} provide eight qualitative multi-turn examples in which a follow-up request changes the target within the same image context.

\end{document}